\documentclass[graybox, envcountchap]{svmult}

\usepackage{imakeidx}         % allows index generation
\usepackage{graphicx}        % standard LaTeX graphics tool
\usepackage{multicol}        % used for the two-column index
\usepackage{multirow}        % used for the two-row index
\usepackage[bottom]{footmisc}% places footnotes at page bottom
\usepackage{tabularx}
\usepackage{colortbl}

\usepackage{newtxtext}       % 
\usepackage[varvw]{newtxmath}       % selects Times Roman as basic font

\usepackage{siunitx}

\makeindex

\begin{document}

\frontmatter%%%%%%%%%%%%%%%%%%%%%%%%%%%%%%%%%%%%%%%%%%%%%%%%%%%%%%

\tableofcontents

\mainmatter%%%%%%%%%%%%%%%%%%%%%%%%%%%%%%%%%%%%%%%%%%%%%%%%%%%%%%%
%\begin{document}

\title{Machine learning-based optimization workflow of the homogeneity of spunbond nonwovens with human validation}
\titlerunning{ML-based optimization of the homogeneity of spunbond nonwovens}
\author{Viny Saajan Victor, Andre Schmei{\ss}er, Heike Leitte and Simone Gramsch}
%\authorrunning{Victor et al.}
\institute{Viny Saajan Victor \at Fraunhofer ITWM, Fraunhofer-Platz 1, 67663 Kaiserslautern, Germany, \email{viny.saajan.victor@itwm.fraunhofer.de}
\and Andre Schmei{\ss}er \at Fraunhofer ITWM, Fraunhofer-Platz 1, 67663 Kaiserslautern, Germany \email{andre.schmeisser@itwm.fraunhofer.de}
\and Heike Leitte \at Technical University of Kaiserslautern, Gottlieb-Daimler-Straße 47, 67663 Kaiserslautern, Germany \email{leitte@cs.uni-kl.de} 
\and Simone Gramsch \at Fraunhofer ITWM, Fraunhofer-Platz 1, 67663 Kaiserslautern, Germany \email{simone.gramsch@itwm.fraunhofer.de}}
\maketitle
\abstract{ In the last ten years, the average annual growth rate of nonwoven production was 4\%. In 2020 and 2021, nonwoven production has increased even further due to the huge demand for nonwoven products needed for protective clothing such as FFP2 masks to combat the COVID19 pandemic. Optimizing the production process is still a challenge due to its high nonlinearity. 
In this paper, we present a machine learning-based optimization workflow aimed at improving the homogeneity of spunbond nonwovens.
The optimization workflow is based on a mathematical model that simulates the microstructures of nonwovens. Based on trainingy data coming from this simulator, different machine learning algorithms are trained in order to find a surrogate model for the time-consuming simulator. Human validation is employed to verify the outputs of machine learning algorithms by assessing the aesthetics of the nonwovens. We include scientific and expert knowledge into the training data to reduce the computational costs involved in the optimization process.  We demonstrate the necessity and effectiveness of our workflow in optimizing the homogeneity of nonwovens.}

% In the last ten years, the average annual growth rate of nonwoven production was 4\%. In 2020 and 2021, nonwoven production has increased even further due to the huge demand for nonwoven products needed for protective clothing such as FFP2 masks to combat the COVID19 pandemic. Optimizing the production process is still a challenge due to its high nonlinearity. 
% In this paper, we present a machine learning-based optimization workflow aimed at improving the homogeneity of spunbond nonwovens. The optimization workflow is based on a mathematical model that simulates the microstructures of nonwovens. Based on training data coming from this simulator, different machine learning algorithms are trained in order to find a surrogate model for the time-consuming simulator. We include expert knowledge to define valid process parameter ranges for specific products. Using the visual parameter space analysis tool presented in ~\cite{victor2022visual}, we demonstrate the application of this optimization workflow to a specific nonwoven product of the company Oerlikon Neumag.

\section{Introduction}
\label{sec:1}
Spunbond processes\index{spunbond processes} are highly effective and cost-efficient methods of producing industrial nonwovens\index{industrial nonwovens} with desirable properties. The resulting nonwovens possess excellent tensile strength and tear resistance, which makes them ideal for applications requiring durability.  Spunbond fabrics also have a consistent structure and thickness, thus rendering them highly desirable for applications that demand uniformity. Due to their ability to allow air and moisture to pass through, these materials are well-suited for use in filtration. Furthermore, spunbond fabrics are an economical alternative to woven or knitted fabrics, making them appropriate for a variety of specialized applications.
These applications include liquid and gas filtration (such as vacuum cleaner bags and water filtration systems), insulation (for roofs, floors, and walls), automotive applications (such as seat covers, door panels, and headliners), medical applications (such as surgical gowns, masks, and drapes), hygiene products (such as diapers, sanitary pads, and wipes), as well as in batteries, fuel cells, and numerous other areas. It is projected that the market for spunbond nonwoven fabrics on a global scale will experience a compound annual growth rate of \SI{3.2}{\percent}~\cite{marketwatch}.

\begin{figure}
    \centering
    \includegraphics[width=\textwidth]{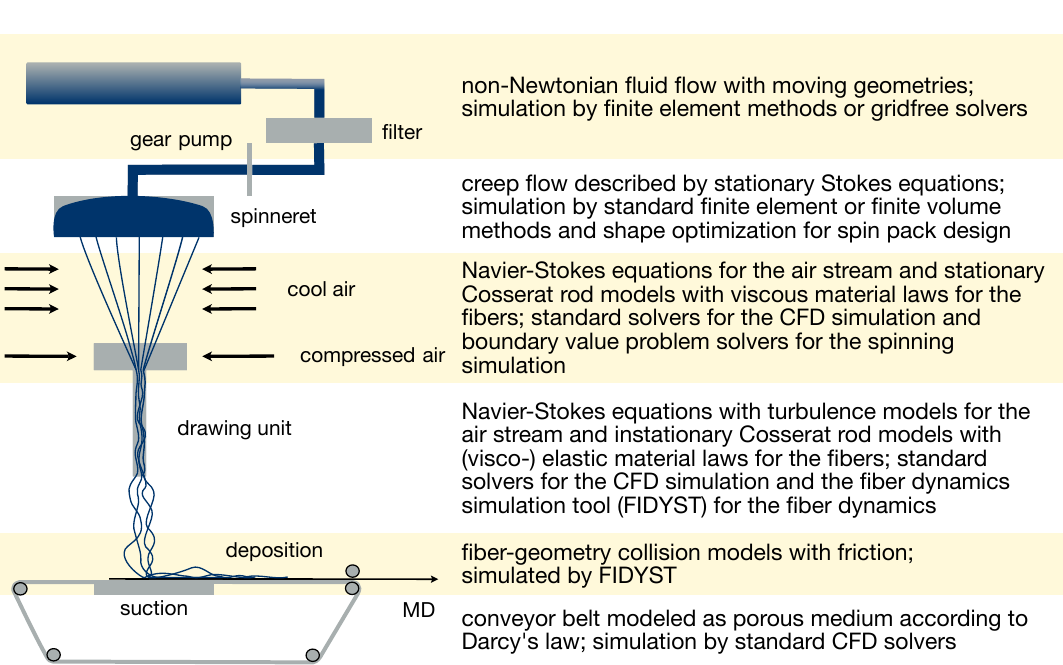}
    \caption{Principle of a spunbond process and necessary simulation models (due to \cite{gramsch:2020}).}
    \label{fig:spunbond_model}
\end{figure}

Using a spunbond process a nonwoven fabric is produced in several steps (cf. Fig.~\ref{fig:spunbond_model}). First, a polymer is melted and pressed through hundreds of nozzles that are positioned in a spinneret. Thereby, filaments are formed from the molten polymer. Air coming from the side cools the polymer fibers. A second air stream stretches the fibers to their final diameter by drawing. After the fibers leave the drawing unit they are twirled by air until they lay down and form the nonwoven fabric. Due to the stochastic nature of the spunbond production process, achieving the desired quality of nonwoven products is a significant challenge for the industry that requires effective control measures. The complex production lines ( $>300$ influencing variables) are typically adjusted by experienced line operators through a trial-and-error process. However, this process fails to reveal the true quality of the line's settings, thus preventing the full utilization of its potential for efficient and sustainable production\index{sustainable production}. As a result, the level of performance, product quality, and line availability remain low. Additionally, the knowledge of how to operate production equipment is usually ingrained in the minds of the operators. This makes it difficult to increase production capacity when demand for certain products suddenly increases. This problem was exemplified in 2020, when the demand for FFP2 masks\index{FFP2 masks} suddenly led to a strong demand for nonwoven products. Despite the surge in demand, production capacities could not be scaled up quickly enough.

The first approach to overcome this problem is to simulate the nonwoven production process. Using the simulation, optimal parameter settings for an altered production process can be searched offline and directly applied when changing the process. Depending on which production process is to be simulated, we need different numerical models\index{numerical models}. Fig.~\ref{fig:spunbond_model} shows the process steps and lists the corresponding physical models that are necessary in order to simulate the spunbond process. These simulations are very accurate and independent of the amount of data available. But the long simulation times prevent a timely prediction of quality when process conditions change as these tools are computationally expensive. 
Additionally, the simulation tools cannot easily incorporate human knowledge\index{human knowledge} and weakly measurable criteria such as aesthetics. 

Another approach is using data-driven machine learning (ML)\index{data-driven machine learning} models as they have gained immense importance in the last two decades. According to a 2018 analysis \cite{fraunhofer:2018}, however, not all potential application industries are developing at the same pace. Regarding the turnover potential of machine learning, the experts rank "Production and Industry 4.0"\index{production and industry 4.0} in 9th place after marketing, consumer electronics, banking, mobility, services, agriculture, and telecommunication. The reason for this is that especially in mechanical engineering often only small amounts of data are collected. Usually, production operates around the clock, 24 hours a day and 7 days a week, which restricts the duration of experimental series. Generating target data at industrial production facilities tends to be expensive and any downtime can result in production losses, leading to reduced revenue.

The goal of this work is therefore to combine simulation models with data-driven machine learning models along with human validation to improve the optimization\index{optimization} of nonwoven production processes. We design a machine learning model to accelerate the mapping of process parameters to nonwoven quality. To address the issue of missing training data, we employed simulation tools. We further incorporate scientific and expert knowledge into this training data making our ML model "Informed". A visualization\index{visualization} tool based on the proposed informed ML model is presented in our work~\cite{victor2022visual}. The tool has been designed to cater to users who are domain experts, material scientists, and textile engineers. The proposed workflow is currently being used and tested by academic simulation experts with offline human validation. The subsequent phase involves implementing the same process for industry use.

%\begin{figure}
%\includegraphics[width=\linewidth]{digital_twin.jpeg}
%\caption{Mathematical model as a digital twin to the physical process of spunbond nonwoven production}
%\label{fig:1}
%\end{figure}

\section{Related Work}
\label{sec:2}
\begin{figure}
\includegraphics[width=\linewidth]{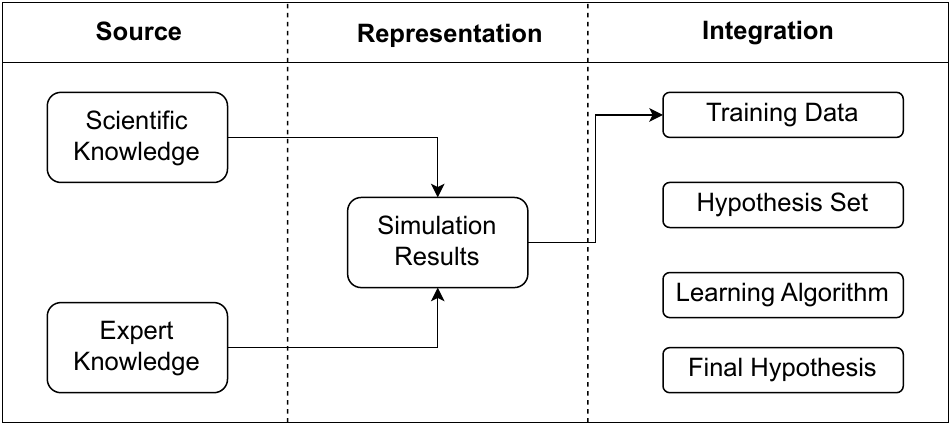}
\caption{Information flow of the knowledge integration in the proposed machine learning method. Diagram adapted from ~\cite{von2019informed}.}
\label{fig:2}
\end{figure}

Recently, data-driven machine learning techniques have produced impressive outcomes due to their ability to recognize patterns and structures in data, allowing for real-time prediction and optimization. However, when confronted with systems that lack sufficient data and demand physical validity, these models are constrained due to their inherent lack of domain expertise. To address this issue, informed machine learning\index{informed machine learning} is used which involves integrating prior, problem-specific knowledge into the machine learning pipeline to improve the system's accuracy and trustworthiness. As presented in ~\cite{von2019informed}, the knowledge can come from various sources and be represented in different forms and injected at various stages of the ML pipeline. Our approach involves incorporating the knowledge derived from simulation results into the training phase of the pipeline. Many previous methods include simulation-based knowledge integration\index{knowledge integration} in the training data by transforming or supplementing input and output features ~\cite{daw2017physics, lee2018spigan, pfrommer2018optimisation, siml, lerer2016learning, rai2019using, shrivastava2017learning}. Our proposed workflow first incorporates expert knowledge\index{expert knowledge} to select relevant features and establish their acceptable ranges for the creation of training data. Following this, scientific knowledge\index{scientific knowledge} is utilized with the aid of simulators to select and validate the input and output features. The presented figure~\ref{fig:2} depicts how the pieces of prior knowledge is represented and integrated into the machine learning workflow.

In the textile industry,\index{textile industry} ML methods are commonly utilized as a substitute model to expedite the optimization process. These models forecast the physical characteristics of the product based on process parameters, allowing for optimization. Various machine learning algorithms are employed to successfully solve classification and regression\index{regression} tasks such as defect detection and quality estimation ~\cite{fan1998worsted, beltran2005predicting, yap2010prediction, abou2015predictingl, eltayib2016prediction, ribeiro2020predicting1, ribeiro2020predicting}. These studies have assessed different ML algorithms to select the most efficient surrogate model for a particular dataset and application. Our study involves training a surrogate ML model to predict product quality from process parameters\index{process parameters}, which is then utilized to create a visualization tool to assist textile engineers in optimizing nonwoven quality. Additionally, our approach incorporates offline human validation of machine-learning model results based on product-specific aesthetics. To the best of our knowledge, no previous study has provided a comprehensive workflow that covers dataset generation to visual application in the context of parameter space exploration to optimize nonwoven quality. Our approach minimizes the time required for optimization, through the utilization of ML, and reduces the need for domain expertise by providing a visual aid to navigate the parameter space. This workflow can be generalized to other applications that seek to optimize product quality by identifying the optimal combination of process parameters.

\section{Machine Learning-based Optimization Workflow using Simulation Models}
\begin{figure}
\includegraphics[width=\linewidth]{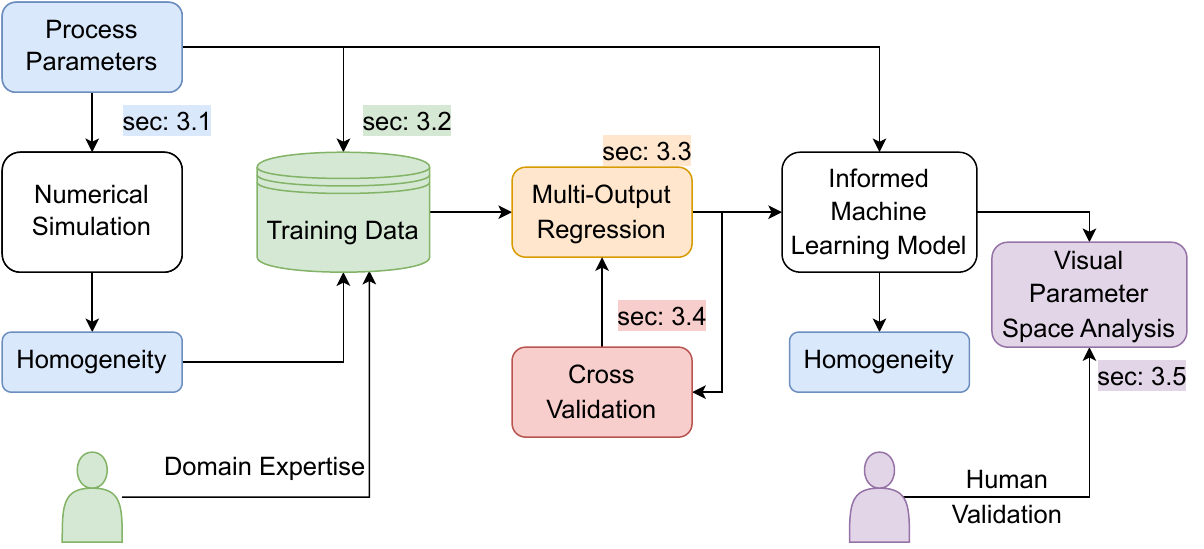}
\caption{Workflow of machine learning-based optimization of spunbond nonwovens. Different stages are colored corresponding to the section numbers in this chapter.}
\label{fig:3}
\end{figure}

% \label{sec:3}
In this section, we propose a workflow for optimizing the quality of spunbond nonwovens based on machine learning. The workflow relies on a numerical tool that simulates the microstructures of nonwoven products using input parameters. However, due to the tool's high computational cost, it is not feasible to utilize it for real-time analysis of nonwoven product quality\index{product quality}. Therefore, we use a machine learning model as a substitute for the tool. The task of the ML model is to predict the quality of the product for varying process conditions. We formulated this ML problem as multi-output regression based on the type of parameters involved in the production process and product quality. The dataset for the ML model is created using the numerical simulation tool as seen in Figure~\ref{fig:3}. 
We integrate scientific and expert knowledge into this dataset. Based on the collected dataset, different regression models are trained and evaluated. The outcomes of the selected ML model are further verified through human validation. Below is a detailed discussion of the workflow that consists of five stages: parameter selection~\ref{sec:1.3.1}, data collection with knowledge integration~\ref{sec:1.3.2}, model selection~\ref{sec:1.3.3}, training and testing~\ref{sec:1.3.4}, homogeneity optimization with human validation~\ref{sec:1.3.5}.

\subsection{Parameter Selection}
\label{sec:1.3.1}
We identified six process parameters that control the quality of spunbond nonwovens. A specific combination of these parameters represents a particular condition involved in the production process. The ranges for the parameters are defined by the domain experts based on typical application scenarios of the final products (for example production of manufacturing material, medical protective masks, etc.)

\subsubsection{Process parameters}
The process parameters correspond to the inputs taken by the numerical simulation tool. They capture the deterministic and stochastic properties of the production process. Each parameter corresponds to a characteristic property. They are chosen such that the characteristic qualities of a real nonwoven and the nonwoven produced by the tool are identical. The selected process parameters are $\sigma_1$, $\sigma_2$, $A$, $v$, $n$, and $d_s$:

\begin{enumerate}
    \item {$\sigma_1$ is the standard deviation of the 2D normal distribution of the fibers around the spinning outlet in the direction which is parallel to the running conveyor belt (machine direction) without belt movement. The feature values vary from \SI{1}{mm} to \SI{50}{mm}.}
    
    \item {$\sigma_2$ is the standard deviation of the 2D normal distribution of the fibers around the spinning outlet in the direction which is perpendicular to the running conveyor belt (cross direction) without belt movement. The feature values vary from \SI{1}{mm} to \SI{50}{mm}.}
    
    \item {$A$ is the noise amplitude of the stochastic process. It is the feature that contains all random effects of the production process, e.g., the influence of the turbulent flow during the fiber spinning, the contacts between fibers, and laydown. This feature specifies whether the simulated fiber lays down in a deterministic (value 0) or stochastic (value $\infty$) manner. The values vary from \num{1} to \num{50}.}

    \item {$v$ is the ratio of spinning speed and belt speed. The feature values vary from \num{0.01} to \num{0.25}.}

    \item {$n$ is the number of spin positions per meter, which can be altered when designing the machine but is fixed during production. The feature values vary from \num{200} to \num{10000}.}

    \item {$d_s$ is the discretization step size which is the distance of discrete points along the simulated fiber curves. The feature values varies from \SI{2.5e-5}{mm} to \SI{5e-5}{mm}}
\end{enumerate}

The selected parameters, apart from $v$ and $n$, do not correspond directly to machine settings, but rather reflect the stochastic properties of the nonwoven produced, which are influenced by various factors such as machine geometry, process parameters (e.g., airflow), process conditions (e.g., air pressure, temperature), and complex interactions (e.g., fiber-fiber, fiber-flow interactions). These parameters are not constant for a machine, except for $n$, and are influenced by the nonwoven material itself. For instance, the density and stiffness of the material affect the curvature of the fibers, which in turn affects the standard deviations $\sigma_1$ and $\sigma_2$. Moreover, these parameters may change depending on the desired application; for instance, producing nonwovens for FFP2 masks requires different operating conditions compared to those for roof materials, as the former must meet strict quality standards. However, changing process conditions in real-time can be challenging because of the down times, such as the time required to clear the old polymer material, refill with new material, and adjust suction position and velocity. To address this issue, the fibers are simulated to depict real-world scenarios, and stochastic properties are assigned to each of them to create virtual nonwoven materials for quality inspection.
\smallskip

\subsubsection{Product Quality : Homogeneity}
The quality of the virtual nonwoven produced from the simulation tools is measured using the coefficient of variation (CV). 
The coefficient of variation is a statistical measure of the relative dispersion of data points in a data series around the mean $\mu$. With the standard deviation $\sigma$ it is defined as 
\begin{equation}
\label{eq:1}
CV := \frac{\sigma}{\mu}
\end{equation}
It establishes the nonwoven web's homogeneity. A more homogeneous nonwoven typically has a lower CV value, which can have an impact on characteristics like filter quality and tensile strength. We compute the CV value at multiple grid resolutions, resulting in an output feature vector (one entry per resolution), to take into consideration homogeneity at various levels of resolution. We regularly discretize the data and compute the fiber mass per bin to obtain the different resolutions. Seven levels of resolution, according to our experiments, had the best agreement with the results of manual inspection.

\subsection{Data Collection with Knowledge Integration}
\label{sec:1.3.2}
This subsection explains the process of collecting data points that are used to train the ML models. We incorporate scientific knowledge and expert knowledge in the data collection process. This reduces the computational time and memory required in the process and facilitates feature selection. The knowledge integration into the data is validated by experimental results. Training on the `informed data' obtained by this approach makes our ML model `informed'.  In the following subsections, we discuss knowledge integration at various stages of the data collection process.

\subsubsection{Sample Size Estimation for Simulation Model Setup}
\label{sec:1.3.2.1}
The computation time and memory required by the numerical simulation tool to simulate virtual nonwoven material increases with the quantity of the material. This makes it not practical to simulate the entire material during data collection. Hence, we decided to simulate a smaller sample of the material that is representative of the whole material in terms of homogeneity. In order to achieve this, we identified two steps: estimating the size of the sample and simulating only the nonwovens that overlap with the material within this sample size. 

The simulation tool is non-deterministic in nature as it produces slightly different results each time when run with the same process parameter setting. Hence, we need to ensure that the selected sample size should have the least statistical uncertainty. Therefore, we created a dataset with \SI{3125} combinations of process parameters using uniform sampling. We simulated this dataset five times with a sample material size of \qtyproduct{5 x 5}{cm}. The statistical uncertainty was quantified as the coefficient of variation of the simulation results for each process parameter setting across five simulation runs. The parameter setting with the maximum uncertainty was further simulated 100 times with three sample sizes: \qtyproduct{5 x 5}{cm}, \qtyproduct{15 x 50}{cm} and \qtyproduct{25 x 50}{cm} for detailed analysis. Table \ref{table:1} shows the coefficient of variation for the three sample sizes. We can observe from the table that uncertainty reduces with an increase in the sample size. Hence, we decided to choose the sample size of \qtyproduct{25 x 50}{cm} and further reduced the uncertainty by sampling each process parameter five times in the data. This averaging of samples reduces the uncertainty by $\sqrt{5}$ according to the central limit theorem.

\begin{figure}
\includegraphics[width=\linewidth]{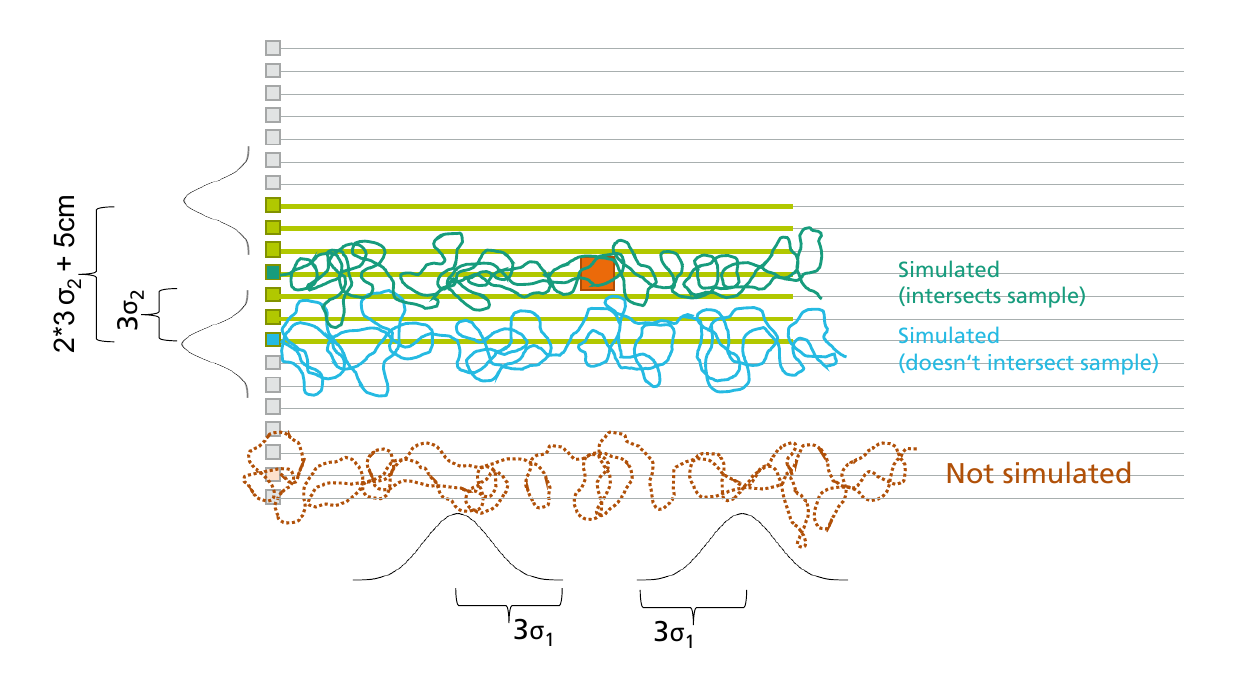}
\caption{Simulation construction based on $\sigma_1$ and $\sigma_2$ values for sample size \qtyproduct{5 x 5}{cm}: Orange area shows the sample extracted for analysis. Green and cyan filaments are simulated in order to compute the sample, brown filament is part of the nonwoven but not simulated.}
\label{fig:4}
\end{figure}

After determining the size of the sample material, we simulate the nonwovens that intersect with this sample material. The laydown of filaments on the conveyor belt is modeled as a 2D normal distribution with standard deviations $\sigma_1$ and $\sigma_2$. According to the empirical rule in statistics, the nonwovens that are a bit more than $3\sigma_2$ away from the sample in the cross direction and a bit more than $3\sigma_1$ away from the sample in the machine direction do not have the chance of overlapping as depicted in Figure~\ref{fig:4}. Hence we do not simulate these filaments.

\begin{table}[!t]
\caption{Table showing coefficient of variation for three sample sizes across seven grid resolutions.}
\label{tab:1}
\begin{tabular}{ p{2cm}p{1.25cm}p{1.25cm}p{1.25cm}p{1.25cm}p{1.25cm}p{1.25cm}p{1.25cm}  }
 \hline
 \multicolumn{8}{c}{Grid resolutions} \\
 \hline
 Sample Size & \SI{0.5}{mm} & \SI{1}{mm} & \SI{2}{mm} & \SI{5}{mm} & \SI{10}{mm} & \SI{20}{mm} & \SI{50}{mm} \\
 \hline
 
\qtyproduct{5 x 5}{cm} & \num{0.04}  & \num{0.05} & \num{0.07} & \num{0.12} & \num{0.21} & \num{0.51}  & \num{0.72}\\

\qtyproduct{15 x 50}{cm} & \num{0.01} & \num{0.01} & \num{0.02} & \num{0.03} & \num{0.05} & \num{0.12}  & \num{0.30}\\

\qtyproduct{25 x 50}{cm} & \num{0.01}  & \num{0.01} & \num{0.01} & \num{0.02} & \num{0.04} & \num{0.09}  & \num{0.20}\\
 \hline
\end{tabular}
\label{table:1}
\end{table}

\subsubsection{Influence of discretization step size ($d_s$)}
\begin{figure}
\includegraphics[width=\linewidth]{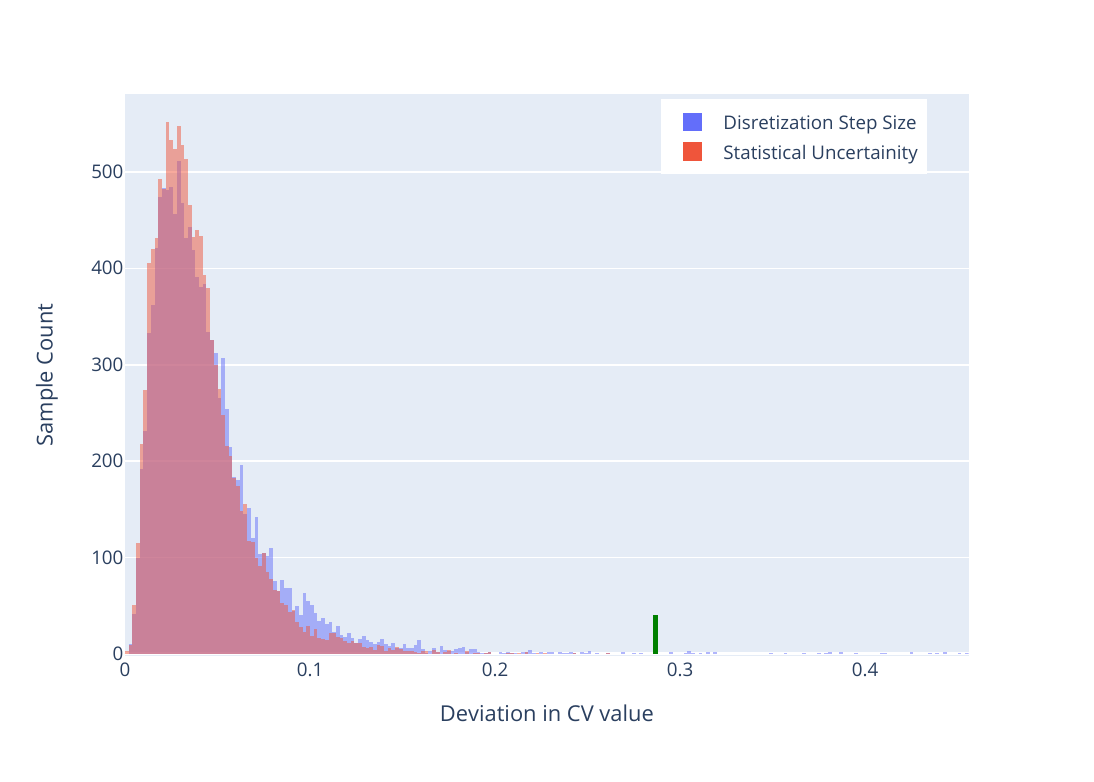}
\caption{Histogram depicting the relative deviation in CV values due to discretization step size (blue) and statistical uncertainty (red).}
\label{fig:5}
\end{figure}

One of the potential parameters that affect the product quality is the discretization step size of the simulated filaments. Based on the domain expertise, it is expected that chosen $d_s$ is small enough such that the simulated nonwoven accurately represents a corresponding real nonwoven and thus this parameter does not affect the simulated product quality. We wanted to investigate whether this assumption is correct. Otherwise, the discretization step size would have to be included as an input parameter in the ML models. For this purpose, we created two small datasets with two different $d_s$ values and analyzed the effect of the two different $d_s$ values on the product quality. Figure~\ref{fig:5} shows the relative deviation in the CV values between two $d_s$ values. The red color denotes the deviation due to the non deterministic statistical uncertainty of the simulation tools and the blue color denotes the deviation due to $d_s$. We defined a threshold to extract the latter alone (green line). We retrieved only \SI{0.0025}{\percent} of the parameter settings that exceeded the threshold. Hence, we concluded that the influence of discretization step size on the product quality is not statistically significant and eliminated this parameter from the input feature set.

\subsubsection{Input Data Sampling}
Once the simulation setup is completed, determining which parameter values to investigate is a crucial question to handle during the data collection. We chose a sparse sampling strategy since the parameter space is very large. Specifically, we used two data sampling techniques to effectively capture the behaviour of the production process for the desired output. The first technique involves the determination of optimal parameter ranges over various small subsets of the parameter space based on the specific nonwoven product. This technique uses expert knowledge to choose the precise parameter ranges. The second technique involves Latin hypercube sampling, which selects parameter values uniformly random from the chosen ranges of values. This technique provides unbiased observations from the numerical simulator.

\subsection{Model Selection}
\label{sec:1.3.3}
We chose multiple-input, multiple-output regression for the machine learning technique since the input data comprises five continuous features ($\sigma_1$, $\sigma_2$, $A$, $v$, $n$) and the output has seven continuous values (CV values at seven grid-sizes). We assessed common regression algorithms such as linear regression, polynomial regression, Bayesian regression,  random forests, and neural networks. A brief description of these models is listed as follows. In order to simplify the optimization formulas described in the following subsections, we have made the assumption that the regression problem involves a single output variable.

\subsubsection{Linear Regression (LR)}
Linear Regression~\cite{montgomery2021introduction} is a supervised machine learning algorithm. It is used to determine the linear relationships between the input parameters and output values. We evaluated four different flavors of the linear regression model: Vanilla, Ridge, Lasso, and ElasticNet. Each of the flavors differs in the type of regularization used in the optimization function. Vanilla regression uses no regularization. Lasso regression and Ridge regression use $L_1$ and $L_2$ regularization respectively. ElasticNet regression uses both $L_1$ and $L_2$ regularization. The minimization problem of the linear regression algorithm with $n$ data points is defined below.
\begin{equation}
\min_{\boldsymbol{w}} \sum_{i=1}^{n} ||\boldsymbol{w}^T \boldsymbol{x_i} - y_i||^2
\end{equation}
where $\boldsymbol{w} = \{ w_1, w_2, ... , w_p \} $ is the coefficient vector and $p$ is the number of input features. The target value $y_i$ is expected to be the linear combination of the i\textsuperscript{th} input feature vector $\boldsymbol{x_i}$. We observed that the linear models are not adequate to fit our data as their error rate is very high. This implies that the relationship between the input parameters and the output values is not linear.

\subsubsection{Support Vector Regression (SVR)}
Support Vector Regression~\cite{smola2004tutorial} is a supervised learning algorithm that is used to predict continuous values. SVR can efficiently perform a non-linear regression using the kernel trick by implicitly mapping the input data into high-dimensional feature spaces. We use the radial basis function as the kernel for our model. The simplest form of the minimization problem for support vector regression is given below.
\begin{equation}
\label{eq:4}
\min \frac{1}{2} ||\boldsymbol{w}||^2
\end{equation}
subject to constraints
\begin{equation*}
\begin{cases}
y_i -  \langle  \; \boldsymbol{w} , \boldsymbol{x_i} \; \rangle - b \leq \epsilon \\
\langle  \; \boldsymbol{w} , \boldsymbol{x_i} \; \rangle + b - y_i \leq \epsilon
\end{cases}
\end{equation*}
where $y_i$ is the target value for the i\textsuperscript{th} input feature $\boldsymbol{x_i}$ and
$\langle  \; , \; \rangle$ denotes the dot product. The goal is to find a hyperplane with optimal values for the  weight vector $\boldsymbol{w}$,
and the bias $b$ that maximizes the width of the margin $\epsilon$ between the predicted outputs and the actual outputs of the training data. SVR uses relatively less memory compared to random forests (with a large number of trees) and artificial neural networks (with a complex architecture) and performs better than linear regression models in high-dimensional spaces. However, it does not scale well with data and is prone to noisy data.

\begin{figure}[b]
\includegraphics[width=\linewidth]{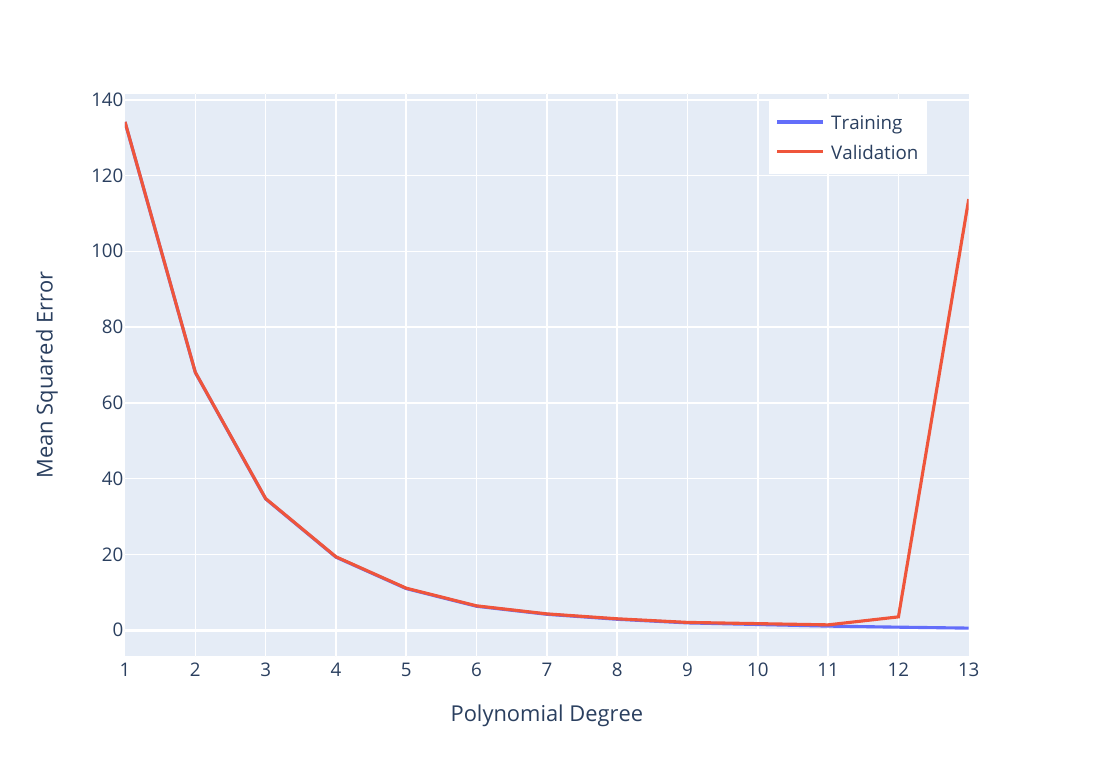}
\caption{Plot showing the mean squared error for different degrees of the polynomial regression.}
\label{fig:6}
\end{figure}

\subsubsection{Polynomial Regression (PR)}
Polynomial regression~\cite{ostertagova2012modelling} is a supervised learning algorithm in which the relationship between the input parameters and the output values is modeled as a polynomial of degree n. The minimization problem for polynomial regression with $n$ data points is defined below.
\begin{equation}
    \label{eq:2}
    \min \sum_{i=1}^{n}(y_i - f(\boldsymbol{x_i}))^2
\end{equation}
where $y_i$ and $f(\boldsymbol{x_i})$ are actual and predicted values  respectively for the i\textsuperscript{th} input feature $\boldsymbol{x_i}$. The polynomial function $f(x)$ can be represented as:
\begin{equation}
    \label{eq:3}
    f(\boldsymbol{x}) = \beta_0 + \beta_1 \boldsymbol{x} + \beta_2 \boldsymbol{x}^2 + ... + \beta_m \boldsymbol{x}^m
\end{equation}
where $\beta_0$, $\beta_1$, $\beta_2$, ..., $\beta_n$ are the coefficients of the polynomial function to be estimated. The degree $m$ of the polynomial that is chosen is crucial in polynomial regression. A very small degree would under-fit the model. As we pursue higher degrees, the training and validation error initially decreases, as seen in Figure~\ref{fig:6}. After degree 11, the training error is still falling while the validation error starts to rise, indicating that the model is beginning to overfit the data. Therefore, we decided that degree 11 would be a good fit for our model. Given that it offers a large variety of functions for data fitting, polynomial regression is well suited to model non-linear relationships between the data. Nevertheless, they are susceptible to overfitting and sensitive to outliers, which has an impact on the generalizability of the models.

\subsubsection{Bayesian Regression (BR)}
In Bayesian regression~\cite{bishop2003bayesian}, problems are formulated using probability distributions rather than point estimates. This enables us to assess the level of uncertainty and confidence in model predictions. The goal of Bayesian 
regression is to ascertain the posterior distribution for the weight vector $\boldsymbol{w}$ rather than to identify the one "best" value. The posterior distribution is the conditional distribution of weight vector $\boldsymbol{w}$ given target variable $y$, hyper-parameter $\alpha$, and model noise variance $\sigma^2$ and it is calculated using Bayesian theorem as given below.
\begin{equation}
    \label{eq:51}
    p(\boldsymbol{w}|y, \alpha, \sigma^2) \propto p(\boldsymbol{w}|\alpha) L(\boldsymbol{w})
\end{equation}
where $p(\boldsymbol{w}|\alpha)$ is the prior distribution of the weight vector which is assumed to be drawn from Gaussian distribution and is given by:
\begin{equation}
    \label{eq:52}
    p(\boldsymbol{w}|\alpha) = (\frac{\alpha}{2\pi})^{1/2} exp\{- \frac{\alpha}{2} ||\boldsymbol{w}||^2\}
\end{equation}
and $L(\boldsymbol{w})$ is the likelihood function which is the conditional distribution of target variable $y$ given weight vector and model noise distribution with mean 0 and variance $\boldsymbol{\sigma}^2$. It is calculated as given below.
\begin{equation}
    \label{eq:53}
    L(\boldsymbol{w}) = p(y|\boldsymbol{w}, \sigma^2) = (\frac{1}{2\pi \sigma^2})^{N/2} exp\{-\frac{1}{2\sigma^2} \sum_{n=1}^{N}|f(\boldsymbol{x_n}; \boldsymbol{w}) - y_n|^2\}
\end{equation}
where $f(\boldsymbol{x_n}; \boldsymbol{w})$ is the function that predicts target variable $y$ for the input feature vector $\boldsymbol{x_n}$ with weight vector $\boldsymbol{w}$.
The goal is to find the value of $\boldsymbol{w}$ that maximizes the posterior distribution which is equivalent to minimizing its negative log. By taking the negative log of the right-hand side of ~\ref{eq:51}, we get the minimization problem of the Bayesian regression as:
\begin{equation}
    \label{eq:54}
    \min \frac{1}{2\sigma^2} \sum_{n=1}^{N} |f(\boldsymbol{x_n}; \boldsymbol{w}) - y_n|^2 + \frac{\alpha}{2} ||\boldsymbol{w}||^2
\end{equation}
The training phase of the Bayesian approach involves optimizing the posterior distribution, and the prediction phase requires additional computation for posterior inference, which involves posterior distribution sampling, log-likelihood computation, and posterior predictive checks. Hence it requires more time for training and inference for larger datasets.

\subsubsection{Random Forests (RF)}
Random Forest Regression~\cite{breiman2001random} is a supervised learning algorithm that leverages the ensemble learning method for regression. This type of learning method combines predictions from various machine learning algorithms to provide predictions that are more accurate than those from a single model. Given the set of input-output pairs ($[\boldsymbol{x},y$]), the goal of random forest regression is to obtain the function $f(\boldsymbol{x})$ the accurately predict the output for unseen input. The function $f(\boldsymbol{x})$ is formulated as an ensemble of $T$ decision trees, where each decision tree $t$ is trained on a subset of training data. The predictions of the decision trees are then combined to obtain final prediction. Each decision tree is built by choosing the feature and threshold that provides the optimal split as determined by a certain criterion (e.g., mean squared error). The data is then divided into subsets based on the chosen feature and threshold. The same steps are repeated for each subset until a stopping criteria is met (e.g., the maximum depth of the tree has been reached ). Finally, a constant value (e.g., mean or median) is assigned to each leaf node as the predicted output value.

Random Forests typically perform well on problems that include features with non-linear correlations. This is supported by their capacity for efficient feature subset selection and rapid decision tree construction, which facilitates faster training and prediction process. However, random forests can be prone to over-fitting, lack interpretability, and suffer from imbalanced datasets.

\subsubsection{Artificial Neural Networks (ANN)}
An artificial neural network~\cite{jain1996artificial} is a computational model that uses a network of functions to comprehend and translate a data input of one form into the desired output. The basic building blocks of a conventional neural network are nodes, which are organized into layers. The input features are passed through these layers (input, hidden, and output) with a sequence of non-linear operations to obtain the final prediction. The minimization problem of neural network for regression with $n$ data points can be formulated as below.
\begin{equation}
    \label{eq:7}
    \min_{ \boldsymbol{W}, \boldsymbol{b}} \sum_{i=1}^{n} L(y_i, f(\boldsymbol{x}_i; \boldsymbol{W}, \boldsymbol{b}))
\end{equation}
where $\boldsymbol{W}$ represents the weights applied to the inputs of a node, while the bias $b$ represents the value added to the weighted sum of the inputs of the same node. and $L$ is the loss function. We used Mean Squared Error (MSE) shown in ~\ref{eq:8} as the loss function for our network. The goal is to find the optimal values for $\boldsymbol{W}$ and $\boldsymbol{b}$ that minimize the loss function. The flexibility of neural networks allows them to learn complex non-linear correlations between inputs and outputs. They can learn to smooth out noise and capture underlying patterns in the data, making them robust to noisy data. However, training neural networks can take a long time, especially if the dataset is large or the model is complex.

We design a Neural Network to learn the non-linear dependency of our output values with respect to the input features. The accuracy of the neural network is determined by the optimal choice of hyper-parameters that decide its architecture. We performed network parameter tuning to find the optimal hyper-parameters while keeping the desirable accuracy. The hyper parameters we used for optimization are the number of hidden layers (from one to five), number of nodes in each layer (from 8 to 1024 with an increment of 8) and the activation functions ('relu', sigmoid' and 'tanh'). We used a randomized search (1000 samples) over the ranges of hyper parameters and selected the parameters based on the error on validation data. Adam optimizer and a learning rate of \num{1e-3} was used for the analysis. We used early stopping method to avoid over-fitting of the data. Table~\ref{tab:2} shows the chosen network architecture based on the evaluation metrics.

\begin{table}[!t]
\caption{Neural network architecture chosen from hyper-parameter tuning.}
\label{tab:2}
\begin{tabular}{p{4cm}p{3.3cm}p{4cm}}
 \hline
 Layer Type & Number of Nodes & Activation Function \\
 \hline
Input Layer & \num{5}  & Linear \\
Hidden Layer & \num{256}  & Relu \\
Hidden Layer & \num{512}  & Relu \\
Hidden Layer & \num{512}  & Relu \\
Hidden Layer & \num{256}  & Relu \\
Hidden Layer & \num{768}  & Relu \\
Output Layer & \num{7}  & Linear \\
 \hline
\end{tabular}
\end{table}

\subsection{Training and Testing}
\label{sec:1.3.4}
We divided the dataset into an \SI{80}{\percent} training data and a \SI{20}{\percent} testing set. The training data is further divided into an \SI{80}{\percent} training set and a \SI{20}{\percent} validation set. As discussed in ~\ref{sec:1.3.2.1}, we sample each data point five times to account for the non-deterministic behavior of the simulation tool. Hence, we divided the dataset into groups of five identical data points and assigned indices to these groups. The indices are then randomly shuffled and split into training, validation, and testing sets based on the proportions described above. This makes sure that the identical data points are assigned entirely to one of the three data sets and the trained ML models can be evaluated for unseen data. The training set is then used to train the regression models and the validation set is used to tune the model hyper-parameters. The testing set is used for unbiased evaluation of the model. Since the input features are measurements of different units, we also performed feature scaling to tailor the data for the machine-learning models. For each data point $x_i$ of the individual input feature distribution $\mathrm{\mathbf{x}}$ having $n$ data points, mean $\mu$, and standard deviation $\sigma$, we calculate the re-scaled feature value $z_i$ as below.
\begin{equation}
z_i := \frac{x_i - \mu}{\sigma}, i \in \{1, 2, \ldots, n\}
\end{equation}

The models are evaluated using the metrics defined below. For the equations used in the following section, we define $\mathrm{\mathbf{Y}}$ and $\mathrm{\mathbf{\hat Y}}$ as the matrices with $m$ rows and $n$ columns, where $m$ is the number of test data points and $n$ is the number of indices corresponding to the seven grid sizes. The individual elements $\mathrm{\mathbf{Y_{i,j}}}$  and $\mathrm{\mathbf{\hat Y_{i,j}}}$ represent the actual and predicted values respectively for the data point $i$ and the grid size corresponding to the index $j$. $\mathrm{\mathbf{\bar Y_{:,j}}}$ represents the mean value of the data points corresponding to the grid size with index $j$.

\begin{enumerate}
    \item {Mean absolute percentage error (MAPE) is a statistical measure to
    evaluate the accuracy of a regression model. The error is independent of the scale of the output as it measures the accuracy as a percentage. MAPE is calculated as below.}

    \begin{equation}
    \label{eq:8}
    MAPE := \frac{100}{n} \sum_{i=1}^{m-1} \sum_{j=0}^{n-1} \left| \frac{\mathrm{\mathbf{Y_{i,j}}}-\mathrm{\mathbf{\hat Y_{i,j}}}}{\mathrm{\mathbf{Y_{i,j}}}}\right| \\
    \end{equation}
    
    \item {Mean squared error (MSE) measures the average of squares of the errors. The MSE is a good estimate for ensuring that the ML model has no outlier predictions with huge errors since it puts larger weight on these errors due to the squaring. MSE is calculated as below.}
    
    \begin{equation}
    \label{eq:9}
    MSE := \frac{1}{n} \sum_{i=1}^{m-1} \sum_{j=0}^{n-1} (\mathrm{\mathbf{Y_{i,j}}}-\mathrm{\mathbf{\hat Y_{i,j}}})^2
    \end{equation}
    
    \item {Coefficient of determination ($R^2$ Score) is the measure of how
    close the data points are to the fitted regression line. It explains how much of the variance of actual data is explained by the predicted values. $R^2$ Score is calculated as below.}

    \begin{equation}
    \label{eq:10}
    R^2(y, \hat y) := 1 -  \frac{\sum_{i=1}^{m-1} \sum_{j=0}^{n-1}  (\mathrm{\mathbf{Y_{i,j}}} - \mathrm{\mathbf{\hat Y_{i,j}}})^2}{\sum_{i=1}^{m-1} \sum_{j=0}^{n-1} (\mathrm{\mathbf{Y_{i,j}}} - \mathrm{\mathbf{\bar Y_{:,j}}})^2}
    \end{equation}
    
\end{enumerate}

\subsection{Homogeneity Optimization with Human Validation}
\label{sec:1.3.5}
\begin{figure}
\includegraphics[width=\linewidth]{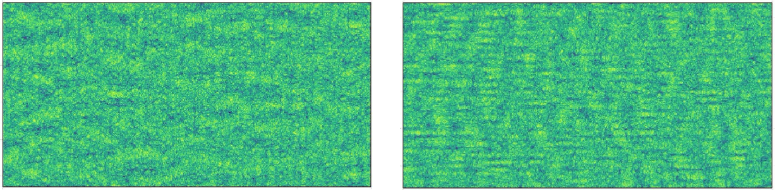}
\caption{Nonwoven materials with same base weight (\SI{2.20}{\gram \per \square\metre}) and similar average homogeneity (\num{13.82} and \num{14.11}), but with different aesthetics.}
\label{fig:7}
\end{figure}

After selecting the best ML model based on the evaluation metrics, it is used to forecast the homogeneity of the spunbond nonwoven given a set of process parameters. However, due to the size of the parameter search space, we are unable to scan the entire parameter space. In order to address this issue, we developed a visualization tool ~\cite{victor2022visual} built on the best ML model that aids textile engineers in parameter space exploration. This tool provides real-time navigation through the parameter space. It also supports the identification of promising regions in the parameter space and the sensitivity of the individual parameter settings. The tool reduces the domain expertise required in the optimization by visually guiding the engineers toward local and global minima. The tool is utilized to identify and choose $n$ potential parameter settings (e.g. $n=10$) based on the optimal CV values. These settings are subsequently used to generate corresponding nonwoven images from the simulator, which is an offline process as it demands a considerable amount of time. Once generated, textile engineers validate the simulated images, discarding any that do not meet the product requirements in order to determine the best parameter setting. One such requirement involves applications where specific aesthetic features (e.g. seat covers for cars, furniture covers) are required. In this case, the best parameter setting can be chosen based on the generated nonwoven's aesthetics. Figure~\ref{fig:7} shows the images of two virtual nonwoven materials with the same base weight (\SI{2.20}{\gram \per \square \metre}) and similar average homogeneity (\num{13.82} and \num{14.11}) with different aesthetics.

\section{Experiments}
\label{sec:4}
In this section, we discuss the necessity and effectiveness of our workflow in optimizing the homogeneity of spunbond nonwovens. For the experiments, we sampled an input database with \num{311740} data points. The dataset included 12348 discrete and 50000 Latin hypercube data points, each of which was sampled five times as discussed in the ~\ref{sec:1.3.2.1}. For each input data point, we simulated the digital nonwoven image using the numerical tool. This image with a selected sample size of \qtyproduct{25 x 50}{cm} is then used to calculate the CV values at seven different grid-sizes. For machine learning analysis, we used  \num{199514} data points for training, \num{49878} data points for validation, and \num{62348} data points for testing.

We evaluated execution times on a workstation with a 40 core Intel\textregistered{} Xeon\textregistered{} E5-2680 v2 (\SI{2.80}{\giga\hertz}) CPU. The execution time to produce \num{311740} virtual spunbond nonwoven samples on this machine is approximately \num{6479983e3} ms with an average time of \num{20786.5} ms per sample.

\begin{table}[!t]
\caption{Performance of machine learning models on the test dataset}
\label{tab:3}
\begin{tabular}{p{4cm}p{2.1cm}p{1.35cm}p{2.1cm}p{1.35cm}p{2.1cm}p{1.35cm}}
\hline
\multirow{2}{*}{ML algorithm}     & \multicolumn{2}{c}{MAPE}              & \multicolumn{2}{c}{MSE}               & \multicolumn{2}{c}{R\textasciicircum{}2 Score} \\ \cline{2-7} 
                                  & \multicolumn{1}{c}{mean}   & variance & \multicolumn{1}{c}{mean}   & variance & \multicolumn{1}{c}{mean}        & variance     \\ \hline
Linear Regression                 & \multicolumn{1}{c}{95.2983} & 1.9254   & \multicolumn{1}{c}{93.2949} & 2.0498   & \multicolumn{1}{c}{0.595}      & \num{2.77e-5}       \\
Support Vector Regression         & \multicolumn{1}{c}{11.0702} & 0.0038   & \multicolumn{1}{c}{22.9310} & 0.4066   & \multicolumn{1}{c}{0.92}      & 0.00      \\
Polynomial Regression & \multicolumn{1}{c}{5.9486} & 0.0093   & \multicolumn{1}{c}{1.767} & 1.1715   & \multicolumn{1}{c}{0.9890}      & \num{1e-5}     \\ 
Bayesian Regression               & \multicolumn{1}{c}{7.7343} & 0.0058   & \multicolumn{1}{c}{1.5080} & 0.0026   & \multicolumn{1}{c}{\textbf{0.99}}      & 0.00       \\ 
Random Forests                    & \multicolumn{1}{c}{10.0065} & 0.0280   & \multicolumn{1}{c}{1.8280} & 0.0137   & \multicolumn{1}{c}{0.98}      & 0.00       \\ 
Artificial Neural Networks        & \multicolumn{1}{c}{\textbf{3.8214}} & 0.0126   & \multicolumn{1}{c}{\textbf{0.358}} & 0.0207   & \multicolumn{1}{c}{\textbf{0.99}}      & 0.00       \\ \hline
\end{tabular}
\end{table}

\begin{figure}
\includegraphics[width=\linewidth]{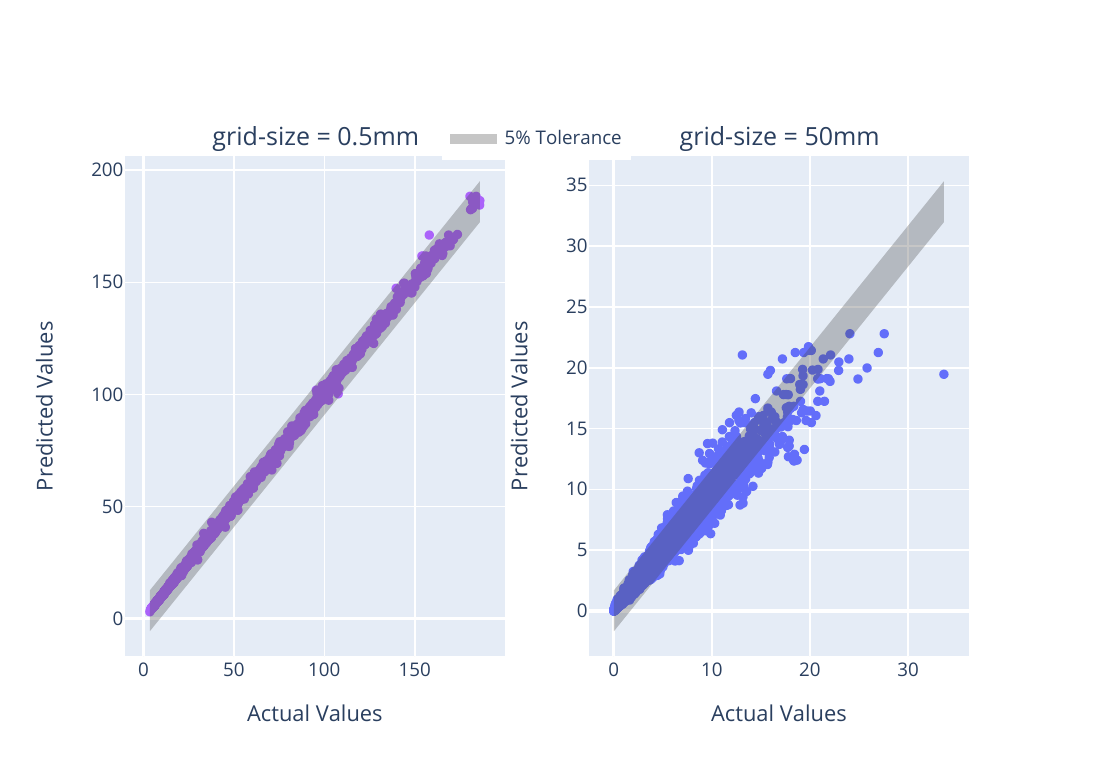}
\caption{Predicted vs actual CV values for grid resolutions \SI{0.5}{mm} and \SI{50}{mm}.}
\label{fig:8}
\end{figure}

\subsection{Models evaluation based on the accuracy}
For statistical evaluation of the ML models, the training and testing sets are randomly selected ten times and for each pair, the models are trained on the training set and evaluated on the testing set. Table~\ref{tab:3} shows the mean and variance of the ten testing set errors for different regression models. We can see from the table that ANNs have the best MAPE, MSE, and $R^2$ Score compared to other models. Figure~\ref{fig:8} shows the CV value predictions versus the actual CV values using ANNs for grid-sizes \SI{0.5}{mm} and \SI{50}{mm}. The grid-size \SI{0.5}{mm} corresponds to the best case with least number of predictions (\SI{6.4031e-3}{\percent} of the testdata) outside the error tolerance range of \SI{5}{\percent}. And the grid-size \SI{50}{mm} corresponds to the worst case with most number of predictions (\SI{1.6712}{\percent} of the testdata) outside the error tolerance range of \SI{5}{\percent}.

\subsection{Models evaluation based on computational performance}
\begin{figure}
\includegraphics[width=\linewidth]{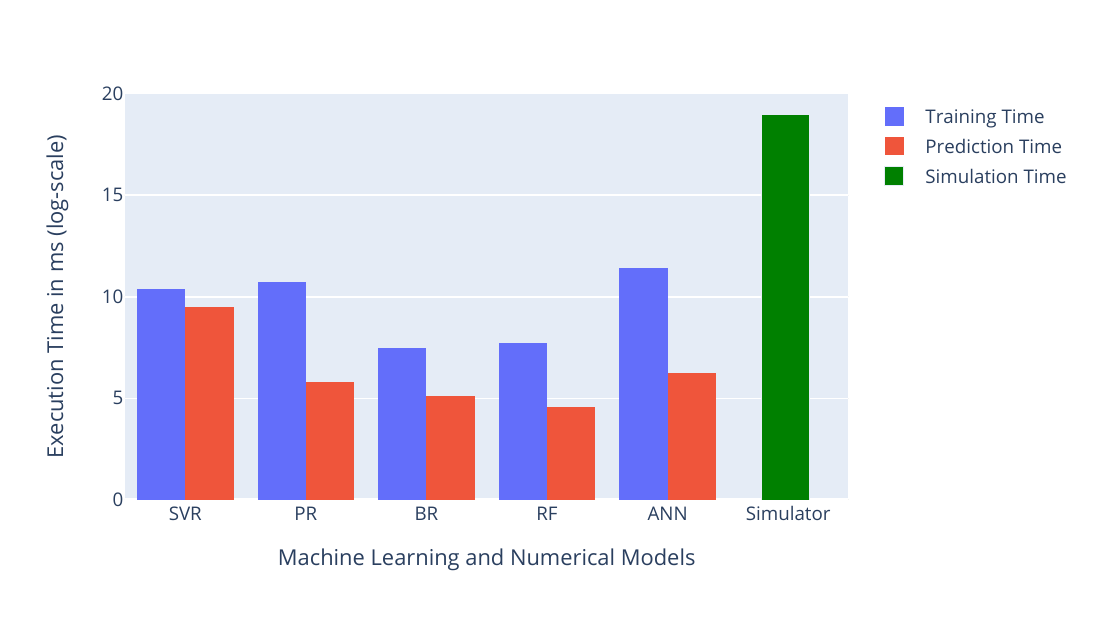}
\caption{Computational performance of ML models compared to the numerical simulator for \num{10000} data samples.}
\label{fig:9}
\end{figure}

The computational efficiencies of the ML models compared to the numerical simulator for \num{10000} data samples are displayed in Figure~\ref{fig:9}. The training time in the figure represents the time required by the model to train on \num{10000} samples, the prediction time shows how long the model requires to forecast \num{10000} samples. The figure shows that Bayesian regression has the shortest training time, and Random Forests have the shortest prediction time.  In general, model training only needs to be done a single time. However, it is necessary to repeat the procedure of employing the models to predict numerous times during optimization. Therefore, Random Forests have a greater advantage relative to other models in terms of computing performance.

\begin{figure}
\includegraphics[width=\linewidth]{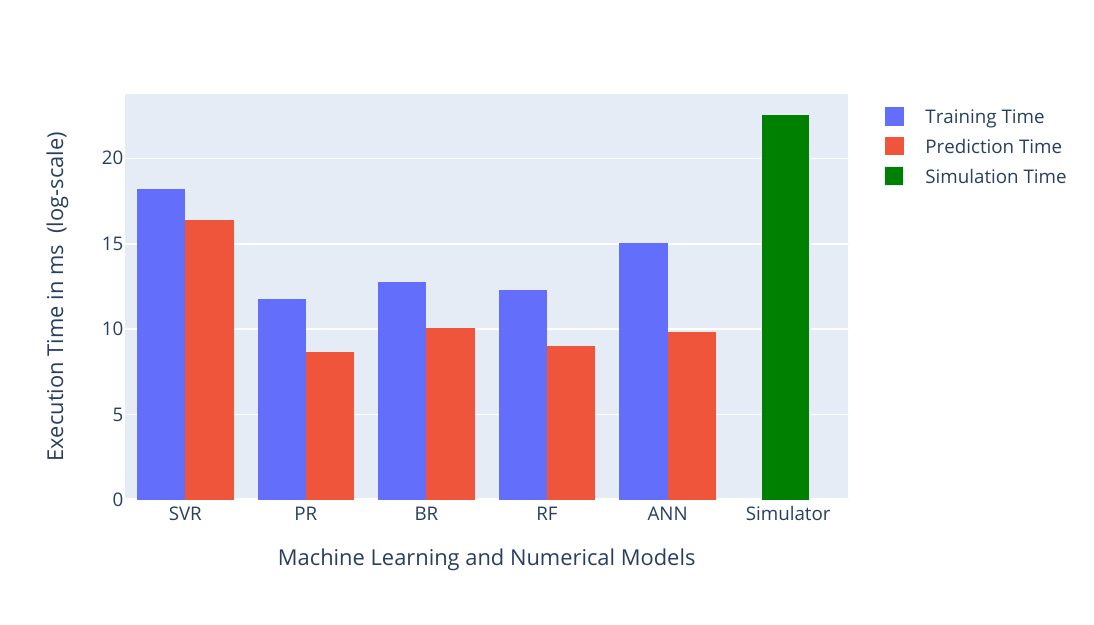}
\caption{Computational performance of ML models compared to the numerical simulator for \num{311740} data samples.}
\label{fig:10}
\end{figure}

For practical applications, the scalability of the ML models plays a significant role. Scalability refers to the ability of ML models to handle large amounts of data and carry out a large number of computations efficiently and quickly.  In order to determine whether the models scale well with the data, we computed the computational time required by the model for a sizable data set. Figure~\ref{fig:10} shows the time taken by the ML models to train and test the entire dataset (\num{311740} samples) in comparison with the numerical simulator. We mainly focus on the prediction time as the training is done only once. The table shows that, with the exception of the SVR, all the models scale reasonably well with the data. The results reveal the effectiveness of using the ML models as a surrogate to the numerical simulation that significantly reduces the time involved in the optimization process (from \num{20786.5} ms to \num{0.0588} ms per sample with ANN as a surrogate). We save approximately \num{74} days for the entire dataset using the ANN model (including the training time).

Based on the evaluation, we chose the artificial neural network as the best surrogate model for the simulation tool as it provided the best accuracy with comparable scalability. This chosen model can be used in real-time optimization of homogeneity of the nonwovens. One successful application of this model is presented in ~\cite{victor2022visual} as discussed before. The proposed visualization tool uses the ANN model for exploring the space of process parameters in real-time to optimize the quality of the nonwovens. The tool is currently tested by academic simulation experts for its efficacy in the optimization process.

\section{Conclusion}
\label{sec:5}
In this chapter, an ML-based workflow for optimizing the homogeneity of spunbond nonwovens is proposed and a model based on multi-output regression is established. During the data collection phase of the training process, we showcased the successful integration of scientific and expert knowledge, leading to the establishment of an Informed ML model. Furthermore, several machine learning algorithms for process parameter tuning are explored based on the model that is verified by human validation. Experimental results show that Artificial Neural Networks have good accuracy and Random Forests have good computational performance across different sizes of training and testing data. Additionally, experimental findings demonstrate the efficacy of our strategy for real-time optimization.

%\end{document}

%%%%%%%%%%%%%%%%%%%%%%%% referenc.tex %%%%%%%%%%%%%%%%%%%%%
% sample references
% 
% Use this file as a template for your own input.
%
%%%%%%%%%%%%%%%%%%%%%%%% Springer%%%%%%%%%%%%%%%%%%%%%%%%%%
%
% BibTeX users please use
% \bibliographystyle{}
% \bibliography{}

%

\backmatter%%%%%%%%%%%%%%%%%%%%%%%%%%%%%%%%%%%%%%%%%%%%%%%%%%%%%%%
\printindex

%%%%%%%%%%%%%%%%%%%%%%%%%%%%%%%%%%%%%%%%%%%%%%%%%%%%%%%%%%%%%%%%%%%%%%

\end{document}